\documentclass[letterpaper, 10 pt, journal, twoside]{IEEEtran}

\usepackage{cite}
\usepackage{graphicx}
\usepackage{picinpar}
\usepackage{amsmath,bm}
\usepackage{url}
\usepackage{colortbl}
\usepackage{soul}
\usepackage{multirow}
\usepackage{pifont}
\usepackage{color}
\usepackage{alltt}
\usepackage[hidelinks]{hyperref}
\usepackage{enumerate}
\usepackage{siunitx}
\usepackage{breakurl}
\usepackage{epstopdf}
\usepackage{pbox}
\usepackage{indentfirst}
\usepackage{booktabs}
\usepackage{makecell}
\usepackage{subfigure}
\usepackage{float}
\usepackage{threeparttable}
\usepackage{algorithmic}
\usepackage{algorithm}
\usepackage{amsfonts}

\def\BibTeX{{\rm B\kern-.05em{\sc i\kern-.025em b}\kern-.08em
    T\kern-.1667em\lower.7ex\hbox{E}\kern-.125emX}}
\usepackage{balance}
\begin{document}
\title{Wheel-SLAM: Simultaneous Localization and Terrain Mapping Using One Wheel-mounted IMU}
\author{Yibin~Wu$^{1,2}$, ~Jian~Kuang$^{1}$, ~Xiaoji~Niu$^{1}$, ~Jens~Behley$^{2}$, ~Lasse~Klingbeil$^{2}$, and~Heiner~Kuhlmann$^{2}$
\thanks{Manuscript received: September 6, 2022; Revised November 4, 2022; Accepted November 28, 2022. This paper was recommended for publication by Editor Javier Civera upon evaluation of the Associate Editor and Reviewers’ comments.}
\thanks{$^{1}$Yibin Wu, Jian Kuang, and Xiaoji Niu are with the GNSS Research Center, Wuhan University, Wuhan, China, \{ybwu, kuang, xjniu\}@whu.edu.cn.}
\thanks{ $^{2}$Yibin Wu, Lasse Klingbeil, Jens Behley, and Heiner Kuhlmann are with the Institute of Geodesy and Geoinformation, University of Bonn, Bonn, Germany, \{firstname.lastname\}@igg.uni-bonn.de.(Corresponding Author: \textit{Jian Kuang})}
\thanks{This work was funded in part by the National Key Research and Development Program of China (No. 2016YFB0501800 and No. 2016YFB0502202) and by the Deutsche Forschungsgemeinschaft (DFG, German Research Foundation) under Germany's Excellence Strategy-EXC 2070-390732324.}}

\markboth{IEEE Robotics and Automation Letters. Preprint Version. November, 2022}{Wu \MakeLowercase{\textit{et al.}}: Wheel-SLAM} 

\maketitle
\begin{abstract}
A reliable pose estimator robust to environmental disturbances is desirable for mobile robots. To this end, inertial measurement units (IMUs) play an important role because they can perceive the full motion state of the vehicle independently. However, it suffers from accumulative error due to inherent noise and bias instability, especially for low-cost sensors. In our previous studies on Wheel-INS \cite{niu2021, wu2021}, we proposed to limit the error drift of the pure inertial navigation system (INS) by mounting an IMU to the wheel of the robot to take advantage of rotation modulation. However, Wheel-INS still drifted over a long period of time due to the lack of external correction signals. In this letter, we propose to exploit the environmental perception ability of Wheel-INS to achieve simultaneous localization and mapping (SLAM) with only one IMU. To be specific, we use the road bank angles (mirrored by the robot roll angles estimated by Wheel-INS) as terrain features to enable the loop closure with a Rao-Blackwellized particle filter. The road bank angle is sampled and stored according to the robot position in the grid maps maintained by the particles. The weights of the particles are updated according to the difference between the currently estimated roll sequence and the terrain map. Field experiments suggest the feasibility of the idea to perform SLAM in Wheel-INS using the robot roll angle estimates. In addition, the positioning accuracy is improved significantly (more than 30\%) over Wheel-INS. The source code of our implementation is publicly available (https://github.com/i2Nav-WHU/Wheel-SLAM).
\end{abstract}


\IEEEpeerreviewmaketitle

\section{Introduction}

\begin{figure}[t]
	\centering
	\includegraphics[width=8cm]{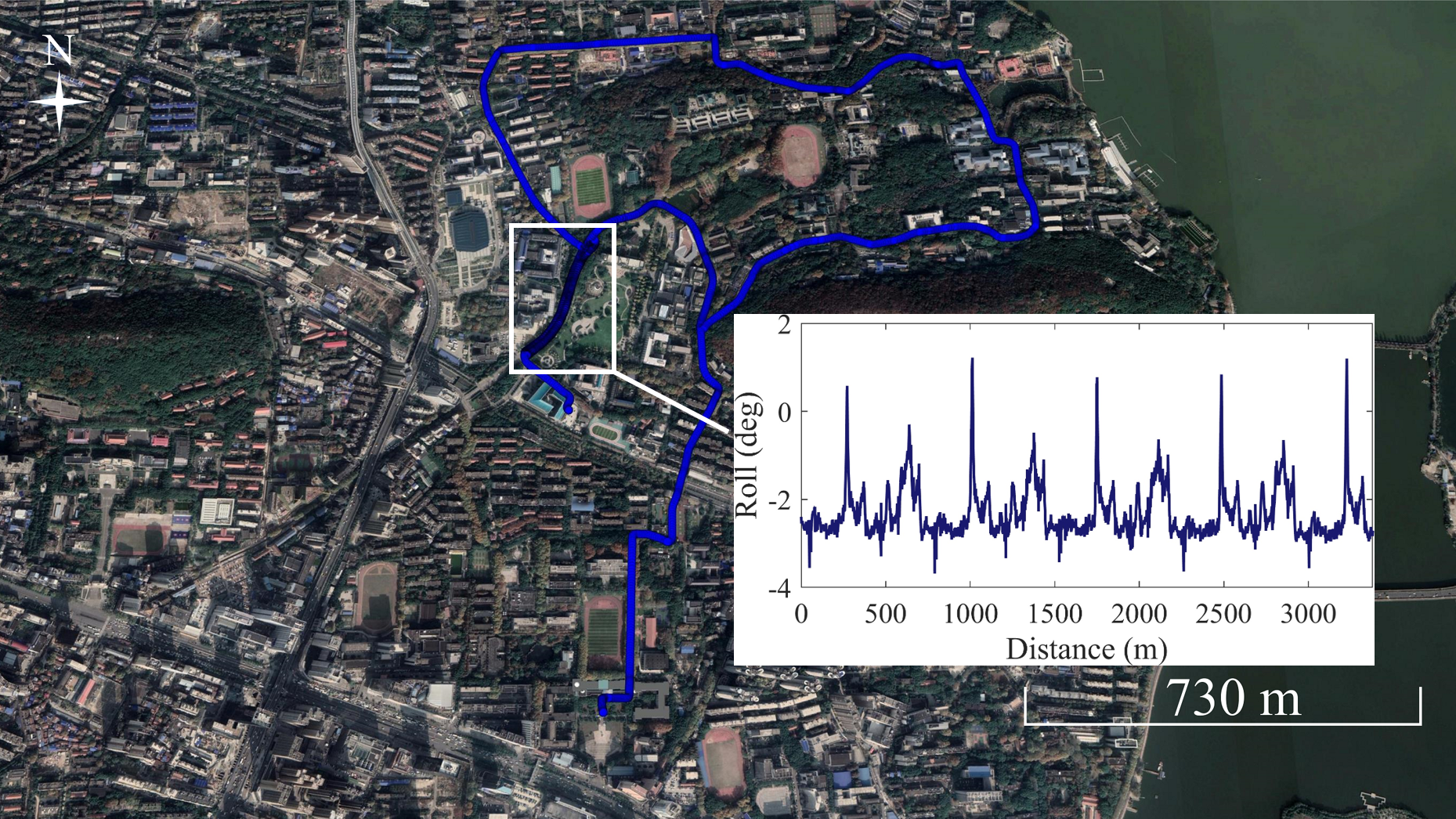}
	\caption{Vehicle roll estimation and test trajectory in the car experiment of our prior work \cite{niu2021}. In the marked area, the car kept circling back and forth. It can be observed that the corresponding robot roll angle estimation (indicating the road bank angle) shows a repeating pattern that can be exploited to perform loop closure detection and correction.}
	\label{examplerollfig}
\end{figure}

\IEEEPARstart{S}{tate} estimation is one of the most fundamental modules for autonomous vehicles. Therefore, modern ground mobile robots are commonly equipped with both exteroceptive sensors, e.g., Global Navigation Satellite System (GNSS) receiver, cameras, Light Detection And Ranging (LiDAR), and proprioceptive sensors, e.g., Inertial Measurement Unit (IMU) and odometer, to track the robot trajectories as well as perceive the environments. Among all these sensors, inertial sensors play a central role in robot navigation due to their self-contained characteristics, which means that it works independently without external signals and interaction with the environment \cite{liyou2021}. In most of the existing works, inertial sensors are used to perform Dead Reckoning (DR) to either complement other navigation techniques \cite{qin2018, huang2019, xu2022TRO} or bridge the signal blockage of other sensors \cite{wu2019sensors}. Although the advances in the microelectromechanical (MEMS) technique have made IMU ubiquitous in various devices, an Inertial Navigation System (INS) suffers from the curse of error drift due to the inherent sensor noise and bias instability. Therefore, it is both challenging and promising to improve the positioning performance of the stand-alone INS.

Inspired by the odometer-aided INS (ODO/INS) \cite{ouyang2020}, we proposed Wheel-INS, a wheel-mounted IMU (Wheel-IMU)-based DR system, in our previous studies \cite{niu2021, wu2021}. Wheel-INS achieved competitive pose estimation performance with only one IMU. It was illustrated that the positioning and heading accuracy of Wheel-INS had been improved by 23\% and 15\% respectively over ODO/INS. Moreover, benefiting from the rotation modulation, Wheel-INS showed desirable resilience to the constant gyro bias error \cite{niu2021}. In summary, there are two major advantages of Wheel-INS. First, a similar information fusion scheme as ODO/INS is achieved by only one inertial sensor. Second, the continuous rotation of the Wheel-IMU significantly limits the heading error drift.  
  
Although Wheel-INS exhibits excellent DR performance, it is only a relative positioning solution lacking the ability to limit long-term error accumulation. That is to say, the positioning error of Wheel-INS still drifts over time, especially when the stochastic error of the inertial sensor is significant. To correct the accumulated error without depending on absolute positioning techniques, a commonly used method is loop detection and relocalization \cite{slamreview2016}. By extracting the environmental features with exteroceptive sensors, e.g., camera and LiDAR, the robot is allowed to recognize places that have been visited and the error drift can consequently be mitigated by, for example, performing a pose graph optimization using loop closure constraints \cite{qin2018}. Then, a natural question arises: \textit{Can we only use the Wheel-IMU to achieve loop closure so as to further limit the error accumulation?} 


From experiments of our prior work \cite{niu2021}, we found that the robot roll estimation in Wheel-INS gave distinguishable and repeatable terrain-correlated responses, as shown in Fig. \ref{examplerollfig}. Therefore, it would be possible to encode the robot roll angle as the road bank feature, so as to enable the loop closure detection in Wheel-INS. In addition, because the IMU is mounted on the wheel which is directly contacted with the ground, the roll angle estimated by Wheel-INS can represent the road bank angle precisely without being affected by the suspension system of the vehicle. 

Based on the discussion above, this letter proposes Wheel-SLAM, a simultaneous localization and terrain mapping system using only one Wheel-IMU. Specifically, we extend our previous study on Wheel-INS \cite{niu2021} by exploiting the robot roll angle estimates to encode the terrain feature which is used for loop closure detection to further limit the error drift of Wheel-INS. Within a particle filter (PF) framework, the uncertainty of the robot state is sampled by multiple particles which maintain their own trajectories and terrain maps. By matching the current roll angle estimation with the map, loop closure can be discovered to update the weights of the particles. In summary, our main contributions include:

\begin{enumerate}[1)]
\item A SLAM system with only one Wheel-IMU using the terrain feature (measured by the Wheel-IMU) is proposed and implemented.

\item We illustrate the feasibility of exploiting robot roll angle estimates to enable loop closure so as to limit the error drift effectively in Wheel-INS through extensive field experiments.

\item To the best of our knowledge, this is the first SLAM system using only one low-cost wheel-mounted IMU for the wheeled robot in the literature. We make our code publicly accessible.
\end{enumerate}

Note that this letter mainly focuses on the feasibility of the idea (achieving loop closure with only one Wheel-IMU in Wheel-INS) instead of the development of a complete system that can be straightforwardly applied to practical applications.  

\section{Related Work}

\subsection{Terrain Matching-based Localization}
Terrain-based vehicle localization provides a usable alternative to GNSS to obtain absolute positioning results by exploiting road information. The repetitive, location-dependent nature of the terrain features allows them to be used for robot localization and mapping \cite{litianyi2019}. Usually, the terrain features are extracted with the in-vehicle inertial sensor, for example, an IMU can measure the road grades by vehicle pitch, road bank angles by vehicle roll, and road curvature by vehicle yaw rate \cite{Zhang2020rs}. The basic hypothesis is that the inertial sensor signals imply vehicle responses to the terrain surface and the same terrain surface excites similar vehicle motions \cite{Gim2021Access}. Existing literature mainly adopts vehicle pitch angles and pitch differences as the features for the terrain-based localization \cite{Vemulapalli2011ACC, Emil2015TITS}, although the roll angle can play the same role \cite{Ahmed2016VTC, Dean2008PhD}. However, using onboard inertial signals to determine the terrain information would be affected by the vehicle maneuver, for example, the braking may induce unexpected pitch variation of the vehicle \cite{litianyi2019} and the centripetal acceleration may introduce the difference between the vehicle roll angle and the road bank angle \cite{Ryu2004ACC}. 



Various methods have been investigated to integrate the terrain feature matching result into the localization pipeline \cite{litianyi2019, Gim2021Access}, while PF has attracted more attention in recent studies. In PF methods, the weights of the particles are updated by evaluating the difference between the in-vehicle measurements (e.g., roll, pitch, etc.) and the response in the map \cite{Dean2011VSD, litianyi2019, Ahmed2016VTC}. Martini et al. \cite{Martini2006PhD} used the Pearson-product correlation coefficient as a distance metric to compare the road grade measurements with the reference map. 



However, a pre-built map is indispensable in all these methods which limits its application. In addition, the inertial sensors used to measure the terrain information are all placed on the vehicle body. By extending our previous studies \cite{niu2021, wu2021}, this letter proposes to localize the robot and measure the road bank angles simultaneously without a prior map. As the IMU is mounted on the robot wheel, the terrain matching would not be affected by the vehicle maneuver which is the case when the IMU is mounted on the vehicle body.

\subsection{SLAM Only Using Inertial Sensors}
Recently deep learning-based inertial localization systems have shown promising results in both pedestrian~\cite{chen2018aaai, herath2022CVPR} and vehicular navigation \cite{martin2020tiv}. These methods learn either motion information from raw inertial measurements \cite{chen2018aaai, herath2022CVPR} or dynamic measurement noise \cite{martin2020tiv} to solve the inertial odometry problem in a data-driven way but fail to exploit environmental signals to limit long-term positional drift.

Angermann et al. \cite{Angermann2012IEEEProc} proposed a pedestrian SLAM system using only foot-mounted IMU (FootSLAM) by taking advantage of human perception and cognition. A dynamic Bayesian network was employed to represent the fact that when a pedestrian walks in a constrained environment, e.g., an office building, he or she relies mainly on visual cues to avoid obstacles and determine accessible areas. Specifically, the algorithm was implemented based on a PF where the weights of the particles were updated by the probability of the pedestrian crossing transitions in a regular 2D grid of adjacent hexagons. After that, a probabilistic transition map implicitly encoding the environmental features that influence the pedestrian's visual impression and intention was constructed.  

Different from FootSLAM where the building layout is implicitly used to perform SLAM by taking advantage of human cognition, Wheel-SLAM uses the Wheel-IMU to explicitly extract the terrain feature for loop closure detection.

In conclusion, Wheel-SLAM borrows ideas from both the terrain matching-based vehicle localization and FootSLAM. In contrast to Wheel-INS \cite{niu2021}, we extend the approach by extracting the road features with the Wheel-IMU to enable loop closure, so as to further limit the error drift. In Wheel-SLAM, we maintain and update the grid map in real-time and detect the loop closure using the DR result from Wheel-INS. After that, we use the roll angle sequence matching results between current estimates and the map to update the weights of the particles for the sake of robustness. 


\section{Methodology}
In this section, a brief introduction to Wheel-INS is first provided. Then we explain the details of the PF-based Wheel-SLAM algorithm, including the dynamic Bayesian model, the construction of the terrain map, and the update of the weights of the particles.

\subsection{Background}
Wheel-INS \cite{niu2021} is the foundation of Wheel-SLAM. It is used to provide the robot odometry and roll angle estimation. There are two major advantages of Wheel-INS. First, the wheel velocity can be calculated by the gyro output and wheel radius enabling the same information fusion as ODO/INS with only one IMU (no other sensors). Second, it can take advantage of rotation modulation to limit the error drift of INS. 

Due to the space limitation, we just outline the algorithm of Wheel-INS here. Please refer to our earlier papers \cite{niu2021, wu2021} for details, e.g., the rotation modulation of the Wheel-IMU, the definition of the misalignment errors, etc.

\begin{figure}[t]
	\centering
	\includegraphics[width=7.5cm]{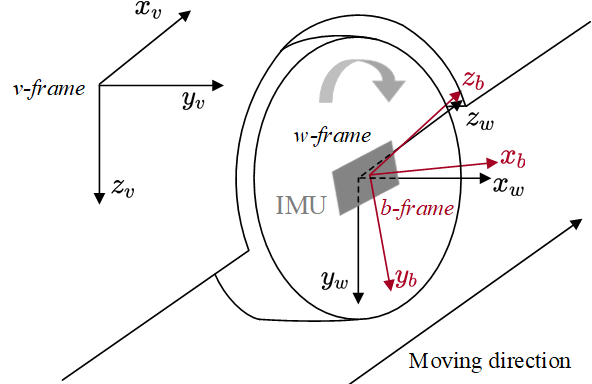}
	\caption{Installation scheme of the Wheel-IMU and the definitions of the vehicle frame (\textit{v}-frame), wheel frame (\textit{w}-frame), and IMU body frame (\textit{b}-frame) \cite{niu2021}.}
	\label{WheelIMUInstallation}
\end{figure}

\begin{figure}[t]
	\centering
	\includegraphics[width=8cm]{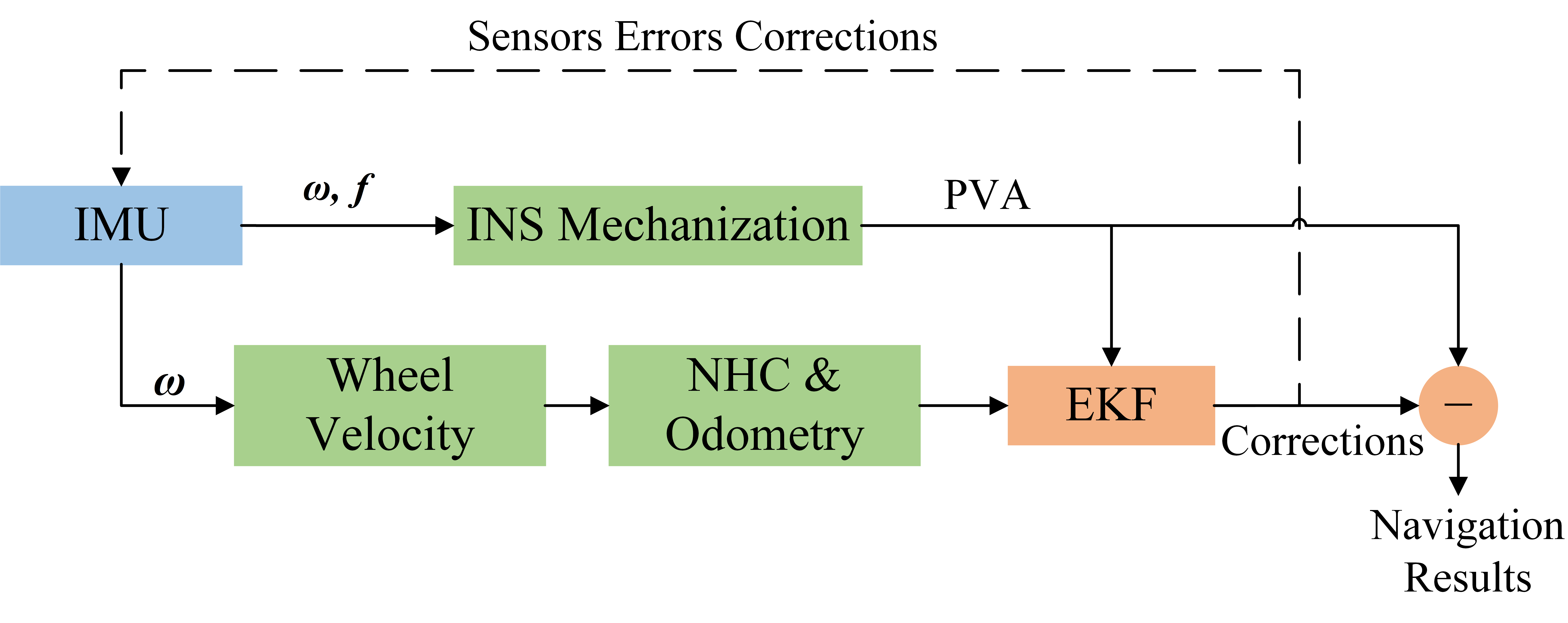}
	\caption{Overview of Wheel-INS \cite{niu2021}. $\bm\omega$ and $\bm{f}$ are the angular rate and specific force measured by the Wheel-IMU, respectively; PVA represents the position, velocity, and attitude of the Wheel-IMU. We use the output from the Wheel-IMU to perform INS mechanization to predict the robot state (PVA). The angular velocity measured in the \textit{x}-axis of the Wheel-IMU and the wheel radius are used to calculate the forward speed. This speed is then integrated with NHC as a 3D velocity observation to update the robot state as well as correct the inertial sensor errors, e.g., gyro bias, through an EKF. }
	\label{WheelINSOverview}
\end{figure}

Fig. \ref{WheelIMUInstallation} depicts the installation of the Wheel-IMU and the definition of related coordinate systems. The system overview of Wheel-INS is shown in Fig. \ref{WheelINSOverview}. First, the forward INS mechanization is performed to predict the robot states. At the same time, the gyroscope outputs in the \textit{x}-axis of the Wheel-IMU are used to calculate the wheel speed. Then, this vehicle velocity is treated as an external observation with non-holonomic constraint (NHC) \cite{dissanayake2001} to update the state through an error-state extended Kalman filter (EKF) \cite{thrun2005probabilistic}. 

The forward wheel velocity calculated by the gyroscope data of the Wheel-IMU and wheel radius can be written as
\begin{equation}
\begin{aligned}
\widetilde{v}^{v}_{wheel} &=\widetilde{\omega}_{x}r-e_v = ({\omega}_{x}+\delta{\omega}_{x})r-e_v\\ &={v}^{v}_{wheel}+r\delta{\omega}_x - e_v
\end{aligned}
\end{equation}
where $\widetilde{v}^{v}_{wheel}$ and ${v}^{v}_{wheel}$ are the observed and true vehicle forward velocity, respectively; $\widetilde{\omega}_{x}$ is the gyroscope output in the \textit{x}-axis; ${\omega}_{x}$ is the true value of the angular rate in the \textit{x}-axis of the Wheel-IMU; $\delta{\omega}_{x}$ is the gyroscope measurement error; $r$ is the wheel radius, and $e_v$ is the observation noise of the wheel velocity, modeled as Gaussian white noise. 

Because the Wheel-IMU rotates with the wheel, the pitch of the robot is unknown in Wheel-INS. In other words, we cannot determine the robot's ascent and descent in Wheel-INS. Therefore, it is assumed that the robot is moving on a horizontal plane. However, experimental results \cite{niu2021} illustrated that this assumption does not cause significant navigation errors. 

\subsection{Dynamic Bayesian Network}
A PF is a sequential Monte Carlo method where the basic idea is the recursive computation of relevant probability distributions using the concepts of importance sampling and approximation of probability distributions with discrete random measures \cite{Djuric2003IEEESPM}. In PF, the posterior distribution of the robot state is represented by a set of particles that evolve recursively with the integration of new information. Based on the technique of Rao-Blackwellization \cite{murphy2001rao, cyrill2007TRO, Montemerlo2002AAAI}, Wheel-SLAM decomposes the SLAM problem into a robot localization problem and a terrain mapping problem that is conditioned on the robot pose estimate. 

In Wheel-SLAM, we try to estimate the posterior
\begin{equation}
p({x}_{1:t},\textbf{M}|{z}_{1:t},{u}_{1:t})
\label{posteriorEq}
\end{equation}
which is the distribution representing the robot state ${x}_{1:t}$ and the terrain map $\textbf{M}$ based on the set of control inputs ${u}_{1:t}$ which governs the robot motion and the road bank angle observations ${z}_{1:t}$. The conditional independence property of the Wheel-SLAM problem implies that the posterior in (\ref{posteriorEq}) can be factored as follows:
\begin{equation}
p({x}_{1:t},\textbf{M}|{z}_{1:t},{u}_{1:t}) = p({x}_{1:t}|{z}_{1:t},{u}_{1:t}) \prod \limits_{i=1}^{N_f} p(m_i|{z}_{1:t},{x}_{1:t})
\end{equation}
where $m_i$ is the \textit{i}-th terrain feature, and $N_f$ is the total number of the features. In Wheel-SLAM, Wheel-INS is performed for the robot state estimation as well as the road bank angle perception.


Wheel-SLAM uses a PF to estimate the robot trajectory distribution. For each particle, the individual trajectory-based terrain map is independent of each other. As a result, every particle is composed of a robot pose and a terrain map; thus, the \textit{i}-th particle at time \textit{t} can be represented as
\begin{equation}
X_i^{t} = \begin{bmatrix}{x}_i^t &\! \textbf{M}_i\end{bmatrix}
\end{equation}
where $i = 1,2,3,...,N_p$, is the index of the particle while ${N_p}$ is the total number of the used particles. $x_i^t$ is the pose of the robot estimated by the \textit{i}-th particle at time \textit{t}, and $\textbf{M}_i$ is the terrain map maintained by the \textit{i}-th particle. Here the robot pose is represented by 2D translation $p_i^t\!\in\!\mathbb{R}^2$ and heading $\textbf{R}_i^t\!\in\!SO(2)$ as Wheel-INS cannot estimate pitch angle, namely, the robot motion in the vertical direction \cite{niu2021}. 

The Wheel-SLAM algorithm consists of four main steps: 1) Sample new robot state by the motion model; 2) Update the terrain map; 3) Update the particle weights once a convinced loop closure is reported; 4) Resample the particles when it is necessary. Algorithm 1 shows an overview of Wheel-SLAM.

\begin{figure}[!t]
	\label{alg1}
	\renewcommand{\algorithmicrequire}{\textbf{Input:}}
	\renewcommand{\algorithmicensure}{\textbf{Output:}}
	\begin{algorithm}[H]
		\caption{Wheel-SLAM}
		\begin{algorithmic}[1]
			\REQUIRE Robot pose $x_{t-1}$, terrain map, $\textbf{M}_{t-1}$, control input, $u_t$ and measurement, $z_t$. 
			\ENSURE Robot pose $x_{t}$, terrain map, $\textbf{M}_{t}$.
            \FOR{each $i=1 \rightarrow N_p$}
                \STATE Predict the vehicle pose by the motion model $p({x}_{t}|{x}_{t-1},{u}_{t})$.
                \STATE Update the terrain map $p(\textbf{M}_{t}|{x}_{t},{z}_{t})$ if necessary.
                \IF {Loop closure detected \&\& Roll sequence matches}   %
            		\STATE Update the weight of the particles according to (\ref{weightupdateEq}).
            	\ENDIF
            \ENDFOR
            \STATE Normalize the weights of the particles $\omega^i = {\omega^i}/{\sum \limits_{i=1}^{N_p} \omega^i}$.  
            \IF {Resampling is required (${N}_\text{eff} = {1}/{\sum \limits_{i=1}^{N_p} (\omega^i)^2} < 0.75$)}          
        		\STATE Perform resampling.
        	\ENDIF
		\end{algorithmic}
	\end{algorithm}
\end{figure}

The first step is to generate a new pose for each particle by sampling from the robot probabilistic motion model:
\begin{equation}
p({x}_{t}|{x}_{t-1},{u}_{t})
\end{equation}
which means that the robot pose, $x_t$, is a probabilistic function of the robot control ${u}_t$ and the previous pose $x_{t-1}$. Here, we adopt Wheel-INS to predict the robot state.

\subsection{Grid Terrain Map Construction}

\begin{figure}[tbp]
	\centering
	\includegraphics[width=7.5cm]{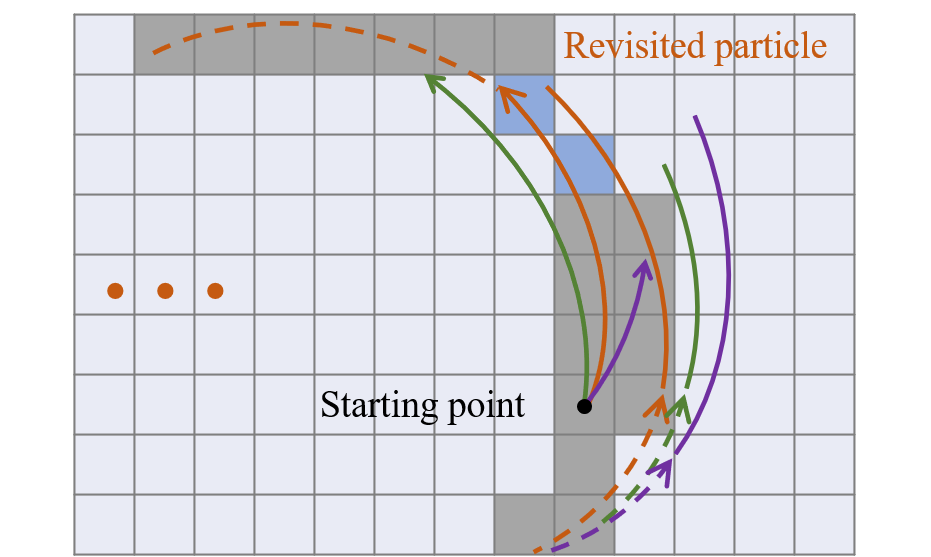}
	\caption{Illustration of the grid map construction and revisit recognition. Different colors of curves represent the robot path sampled by different particles. The gray grids have been visited by the robot thus they have a road bank angle estimation. Once a particle detects that the robot has continuously returned to visited grids (blue), a potential loop closure is reported for further check (Please refer to Section III-D for details).}
	\label{gridmap}
\end{figure}

Compared to the hexagon grid map used in FootSLAM \cite{Angermann2012IEEEProc}, we simplify the grid to square due to the relatively simple robot motion mode. As we assume that the vehicle is moving on the horizontal plane, we build a 2D grid map. Each grid holds the corresponding road bank angle estimated by Wheel-INS at that position. Fig. \ref{gridmap} illustrates the grid map constructed along with the robot pose evolution.

\subsection{Particle Weight Update}
In the beginning, all the particles are assigned the same weight. When the robot moves, each particle has a different trajectory and terrain map. To ensure the reliability of loop closure and reduce the impact of an outlier, we set three criteria. First, loop closures need to be continuously detected by the robot position in a window of length $N_r$. Second, we calculate the $N_r$ roll sequence matching scores using Pearson correlation coefficient \cite{Martini2006PhD} and compare them with a threshold $C_\text{thr}$. In this $N_r$ window, at least $N_\text{thr}$ ($N_\text{thr} <= N_r$) coefficients need to be larger than $C_\text{thr}$. Third, the correlation coefficient at the current position needs to be larger than the threshold. If all three requirements are met, we think it is a true loop closure and subsequently update the particle weights as follows:


\begin{equation}
{\omega}_k^{i} = {\omega}_{k-1}^{i}\frac{N_c}{N_r}\exp(\mathrm{RMSE}(V_\text{coeff}))
\label{weightupdateEq}
\end{equation}
where $k$ indicates the time epoch; $i$ indicates the number of the particle; $N_c$ ($N_c\!\geq\! N_\text{thr}$) is the number of the correlation coefficients larger than the threshold in the window $N_r$; RMSE indicates the root mean square error; $V_\text{coeff}$ is the collection of the correlation coefficients larger than the threshold in the window. After that, normalization is performed to make the sum of the weights equal to one. 

Note that although Wheel-SLAM requires the exact revisit of the robot, the vehicle doesn't need to always drive on the same road. The position error will accumulate when the robot explores an unknown environment, but it can be corrected once the robot is back on a previously-visited road. Please refer to Section IV for the experimental results and discussion. Moreover, as our algorithm requires a sequence of road bank angles for matching, it is not able to detect loop closure effectively at crossroads where the robot may enter a place perpendicular to its last entrance direction.

\section{Experimental Results}
This section presents and discusses the real-world experimental results to illustrate the positioning performance of Wheel-SLAM. As this work focus on the feasibility of the idea (achieving loop closure with only one Wheel-IMU) instead of the development of a complete system, we do not evaluate the computational efficiency of Wheel-SLAM. In addition, there is no paper in the literature studying SLAM with the same sensor set (wheel-mounted inertial sensors). We only compare the proposed Wheel-SLAM with its predecessor, Wheel-INS, to illustrate its effectiveness and discuss its characteristics. First, the experimental conditions are described. After that, we compare the positioning accuracy of Wheel-SLAM with Wheel-INS. The characteristics of Wheel-SLAM are also analyzed. Finally, we further discuss the key components which play a central role in Wheel-SLAM and some limitations in real-world applications. 

\subsection{Experimental Description}
\begin{figure}[t]
	\centering
	\includegraphics[width=8cm]{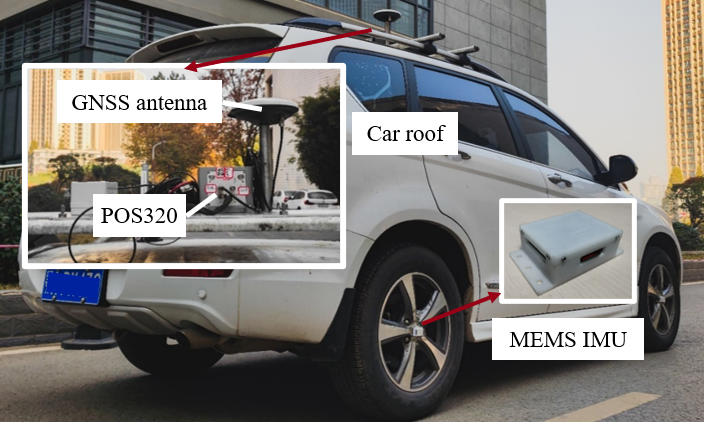}
	\caption{Experimental platform. A GNSS antenna and a high-end IMU (POS320) were mounted on the car roof to provide the ground truth of the vehicle pose while a low-cost MEMS IMU was mounted on the right rear wheel to perform Wheel-SLAM.}
	\label{figexpplatform}
\end{figure}

To demonstrate the feasibility and effectiveness of the proposed Wheel-SLAM system, we carried out five sets of field tests using a car on the campus of Wuhan University. The car was instrumented with one Wheel-IMU and reference system to provide the ground truth of the vehicle pose, as shown in Fig. \ref{figexpplatform}. The characteristics of the vehicle motion in the tests are shown in Table \ref{Tabvehiclemotion}. 

\begin{table}[t]
	\centering
	\caption{Vehicle Motion Information in the Experiments}
	\label{Tabvehiclemotion}
	\begin{tabular}{cccccc}
		\toprule
		{Sequence} & I & II & III  & IV & V \\
		\midrule
		\specialrule{0em}{2pt}{2pt}
		
		\makecell{Average \\ speed ($m/s$)} & \makecell{5.41} & \makecell{5.47} & \makecell{4.93} & \makecell{5.16} & \makecell{4.89} \\ 
		\makecell{Total \\ distance ($m$)} & \makecell{$\approx$2950 } & \makecell{$\approx$3791 } & \makecell{$\approx$2802 } & \makecell{$\approx$7235 } & \makecell{$\approx$9285 } \\
		
		\bottomrule
	\end{tabular}   
\end{table}

\begin{table}[t]
	\centering
	\caption{Technical Parameters of the IMUs Used in the Tests}
	\label{TabIMUparas}
	\begin{threeparttable}
		\begin{tabular}{m{1.2cm}<{\centering}m{1.3cm}<{\centering}m{1.3cm}<{\centering}m{1.3cm}<{\centering}m{1.3cm}<{\centering}}
			\toprule
			IMU & Gyro Bias\newline ($^\circ/h$) & {ARW \newline ($^\circ/\sqrt{h}$)}  & {Acc. Bias \newline ($m/s^2$)} & {VRW \newline ($m/s/\sqrt{h}$)} \\
			\midrule
			\specialrule{0em}{3pt}{3pt}
			POS320 & 0.5  & 0.05 & 0.00025 & 0.1\\
			\specialrule{0em}{3pt}{3pt}
			ICM20602 & 200 & 0.24 & 0.01 & 3\\
			\bottomrule
		\end{tabular}
	\end{threeparttable}
\end{table}

\begin{figure}[t]
	\centering
	\includegraphics[width=8cm]{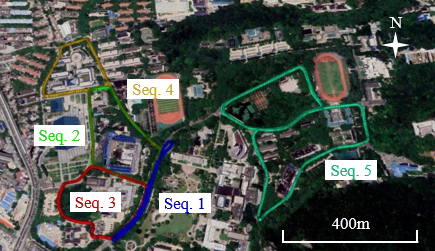}
	\caption{Experimental trajectories. Seq. 1, Seq. 2, and Seq. 3 are circular trajectories where the vehicle moved several times in one direction while Seq. 4 and Seq. 5 are more complicated ones where the vehicle moved in different directions (not just turn around in one direction) in large-scale environments.}
	\label{figtraj}
\end{figure}

We used the same MEMS IMU from our prior works \cite{niu2021, wu2021}. The MEMS IMU contained four ICM20602 (TDK InvenSense) inertial sensor chips, a chargeable battery module, a microprocessor, an SD card for data collection, and a Bluetooth module for communication and data transmission. One can use an android phone to control the data collection. We collected the data of one chip of the Wheel-IMU for post-processing. The reference system used in the experiments was a high-accuracy position and orientation system with a tactical-grade IMU (POS320, MAP Space Time Navigation Technology Co., Ltd., China). The reference data were processed through a smoothed post-processed kinematic (PPK)/INS integration method. The main technical parameters of both the MEMS IMU and the high-end IMU are listed in Table \ref{TabIMUparas}, where ARW denotes the angle random walk; Acc. denotes the accelerometer; VRW denotes the velocity random walk.

Fig. \ref{figtraj} shows the five experimental trajectories. In Sequence (Seq.) 1, the car was traveling back and forth on two parallel and opposite-direction roads. Seq. 2, and Seq. 3 are circular trajectories where the vehicle moved several times in one direction. Seq. 4 and Seq. 5 are more complicated ones with two circles in large-scale environments. In Seq. 4 and Seq. 5, the car not only traveled in the same lane in the same direction but moved in opposite directions on the same road. Note that in Seq. 1, Seq. 2, and Seq. 3, the car sometimes also changed the lane. Please refer to our source code webpage for the street views of the sequences.

The static IMU data before the car started moving were used to obtain the initial roll and pitch angle of the Wheel-IMU, as well as the initial value of the gyroscope bias. Other inertial sensor errors were set as zero. The key parameters of Wheel-SLAM set in the experiments are listed in Table \ref{TabWSpara}. Standard deviation is denoted as STD in Table \ref{TabWSpara}.

\begin{table}[t]
	\centering
	\caption{Key parameters of Wheel-SLAM in the experiments.}
	\label{TabWSpara}
	\begin{threeparttable}
		\begin{tabular}{p{5cm}p{3cm}}
			\toprule
			{Parameters} &  {Value} \\
			\specialrule{0em}{1.5pt}{1.5pt}
			
			\midrule
			\specialrule{0em}{1.5pt}{1.5pt}
			Particle number & 100 \\
			\specialrule{0em}{1.5pt}{1.5pt}
			Grid size of the 2D terrain map & 1.5\,m \\
			\specialrule{0em}{1.5pt}{1.5pt}
			Distance increment STD & 0.025\,m \\ 
			\specialrule{0em}{1.5pt}{1.5pt}
			Heading increment STD & 0.05\,$^\circ$ \\
			\specialrule{0em}{1.5pt}{1.5pt}
			Roll sample distance & 0.5\,m \\
			\specialrule{0em}{1.5pt}{1.5pt}
			Roll matching sequence length & 25\,m \\
			\specialrule{0em}{1.5pt}{1.5pt}
			Correlation coefficient threshold & 0.4 \\
			
			\bottomrule
		\end{tabular}
	\end{threeparttable}
\end{table}

\begin{figure}[!t]
	\begin{tabular}{cc}
		\begin{minipage}[t]{0.5\linewidth}
		    \centering
			\subfigure[Estimated trajectories against ground truth in Seq. 1.]{
			\includegraphics[height = 0.7\linewidth]{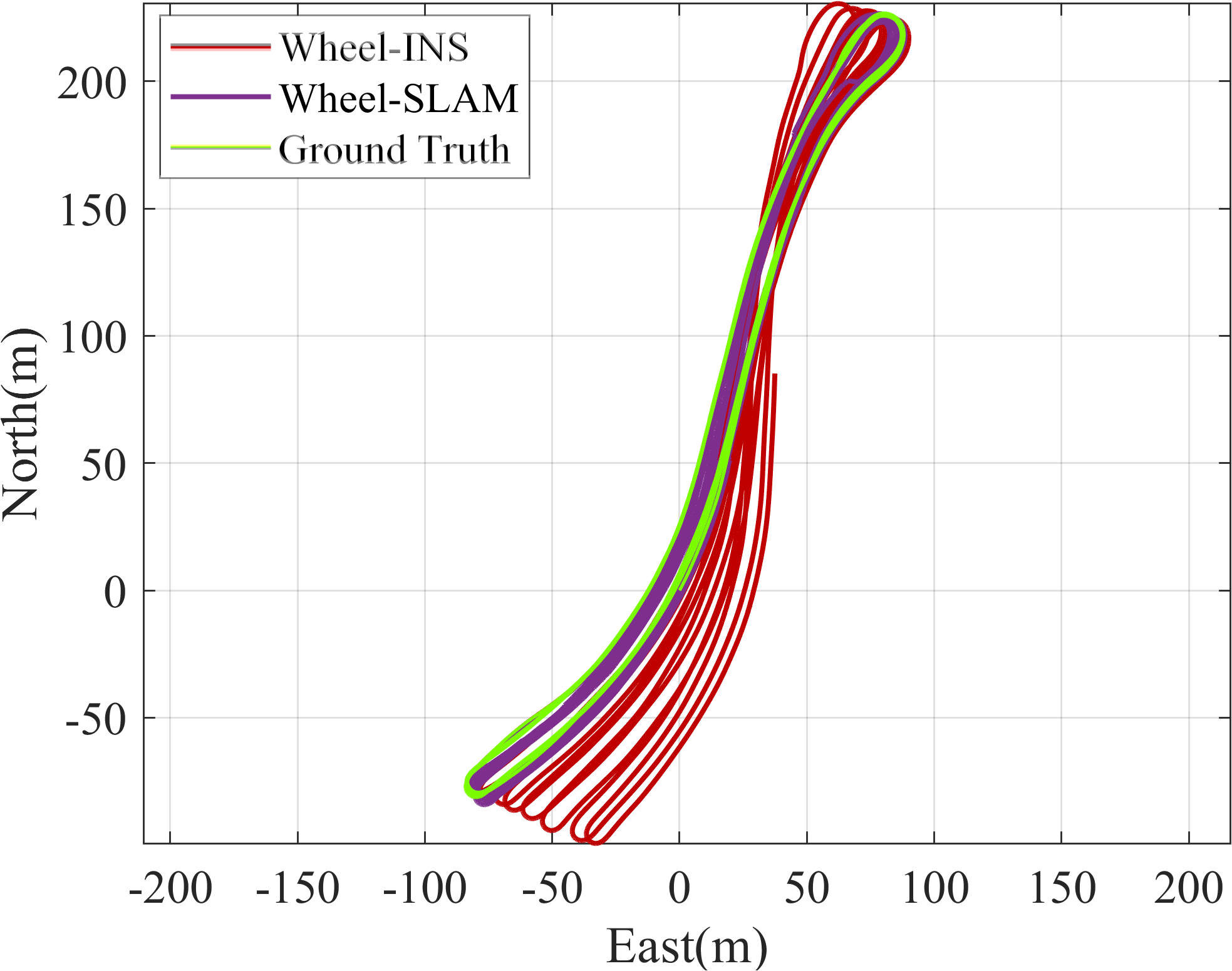}
		}
		\end{minipage}
		\begin{minipage}[t]{0.5\linewidth}
		    \centering
			\subfigure[Horizontal positioning and heading errors in Seq. 1.]{
			\includegraphics[height = 0.7\linewidth]{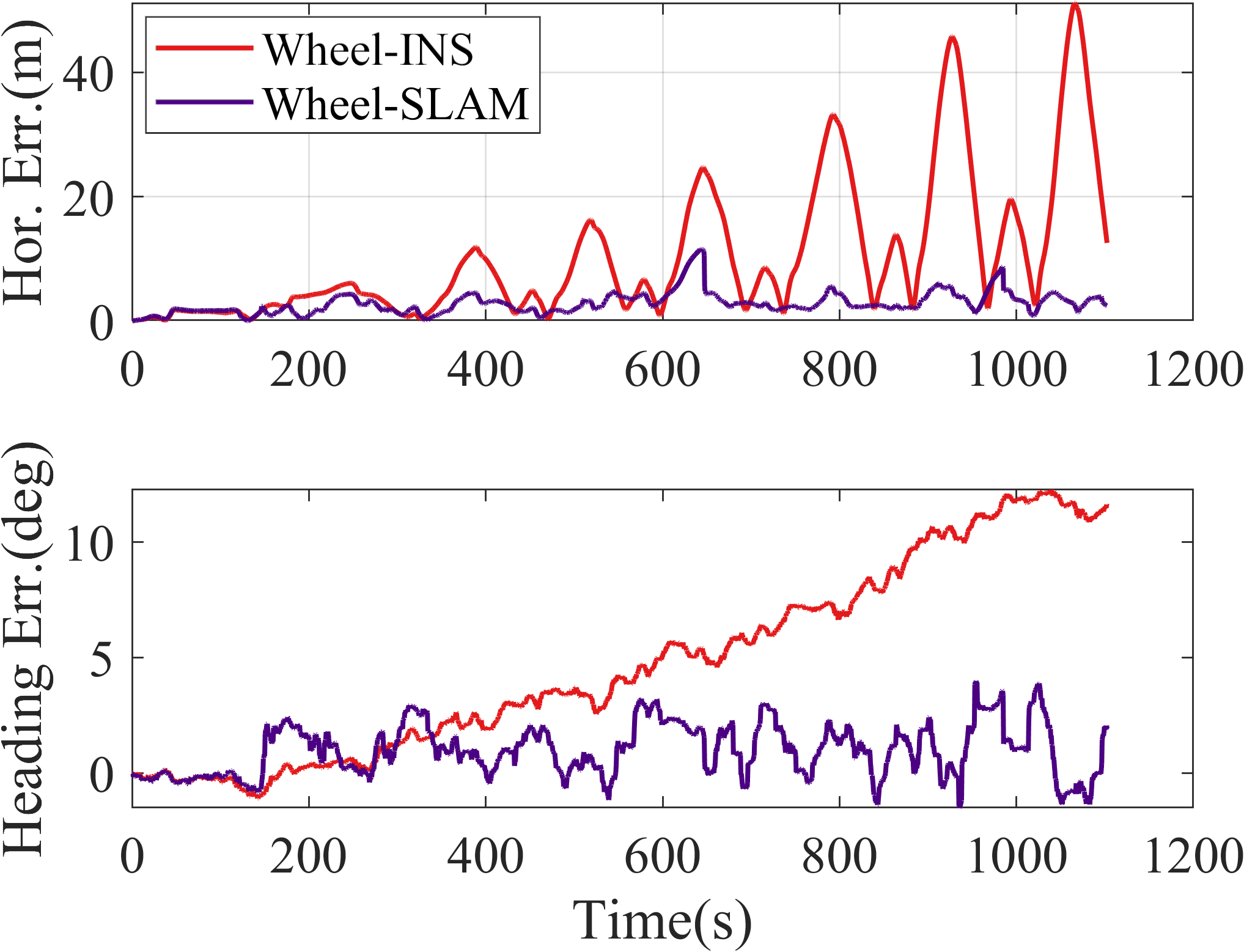}
		}
		\end{minipage}
	\end{tabular}
	
	\begin{tabular}{cc}
		\begin{minipage}[t]{0.5\linewidth}
		    \centering
			\subfigure[Estimated trajectories against ground truth in Seq. 2.]{
				\includegraphics[height = 0.7\linewidth]{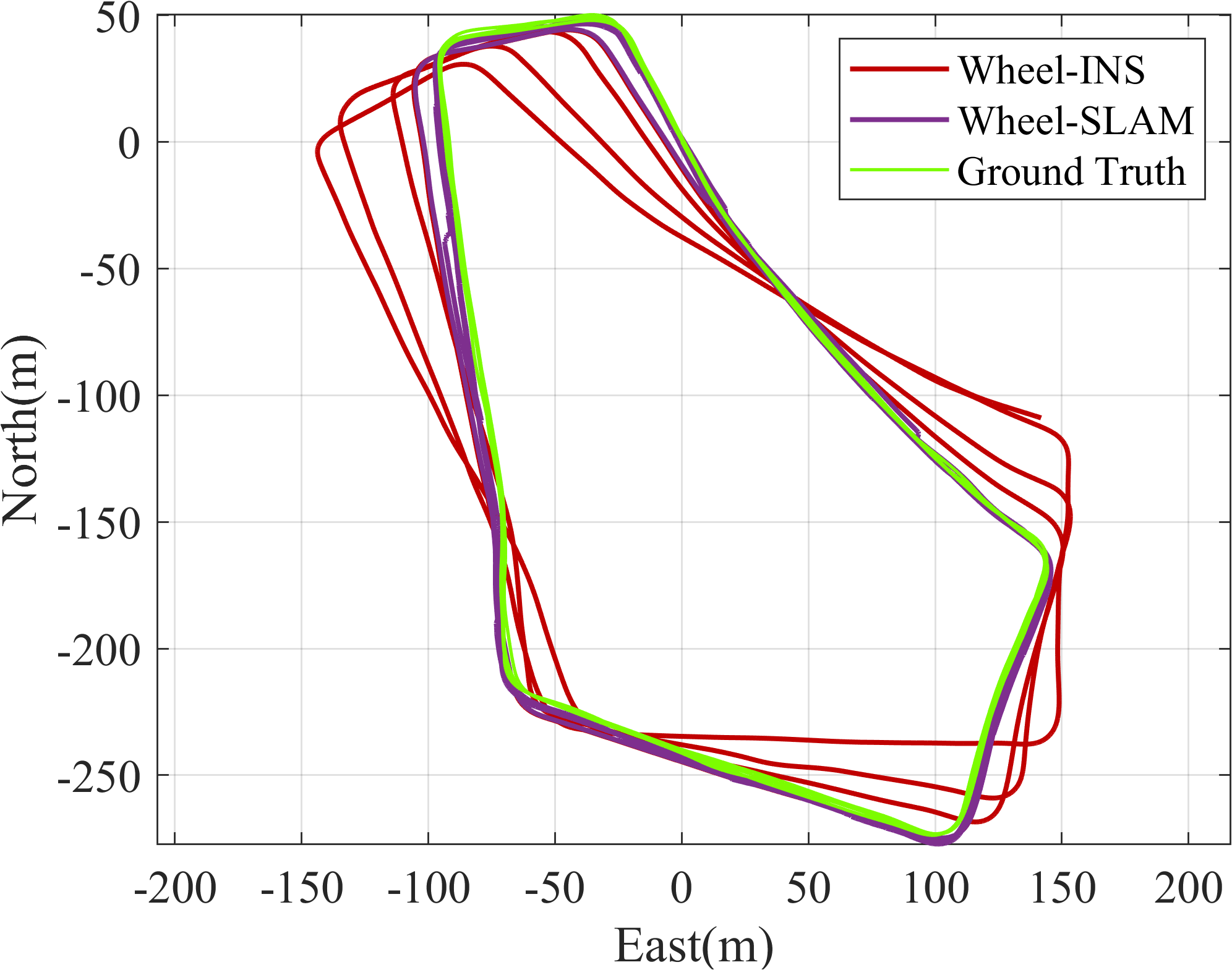}
			}
		\end{minipage}
		\begin{minipage}[t]{0.5\linewidth}
		    \centering
			\subfigure[Horizontal positioning and heading errors in Seq. 2.]{
				\includegraphics[height = 0.7\linewidth]{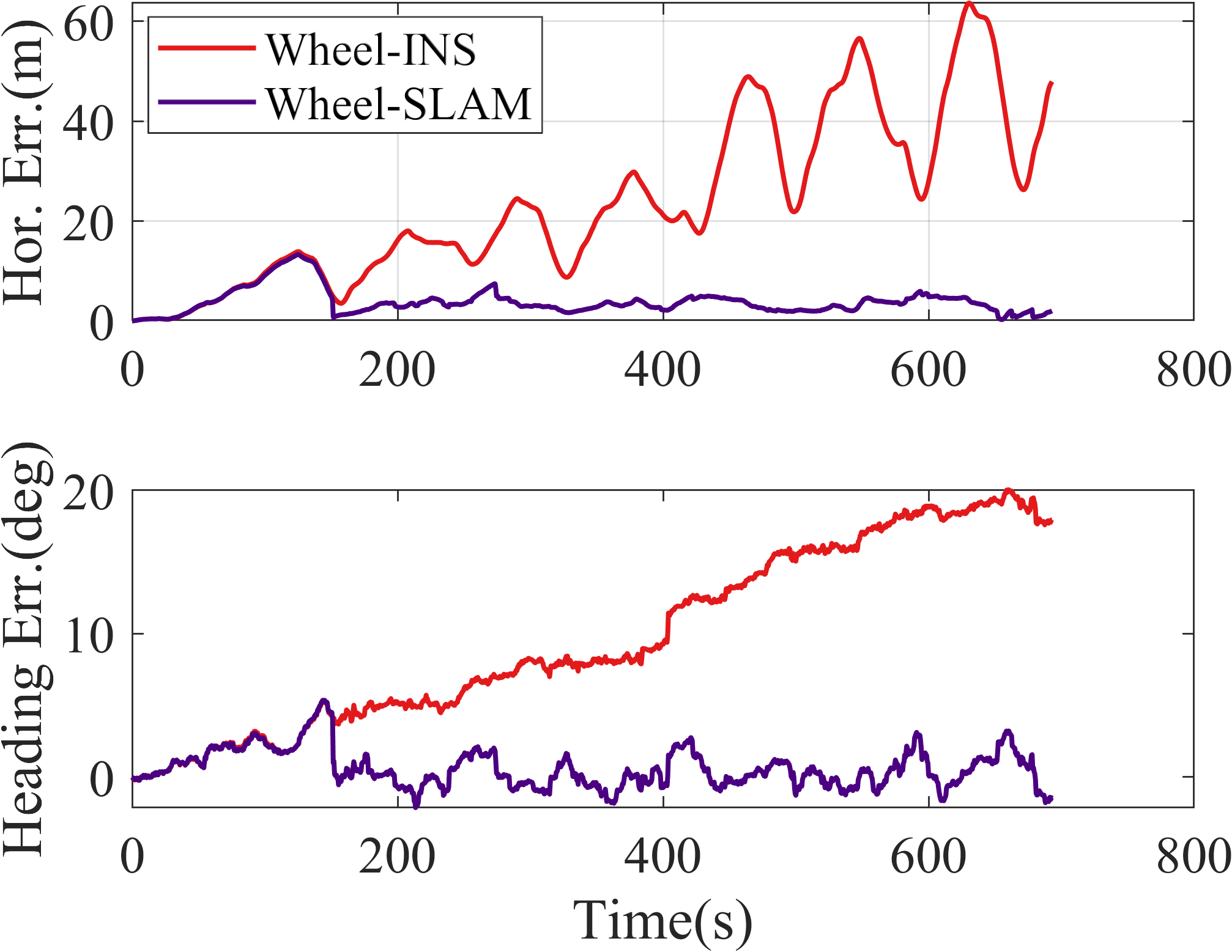}
			}
		\end{minipage}
	\end{tabular}
		
	\begin{tabular}{cc}
		\begin{minipage}[t]{0.5\linewidth}
		    \centering
			\subfigure[Estimated trajectories against ground truth in Seq. 3.]{
				\includegraphics[height = 0.7\linewidth]{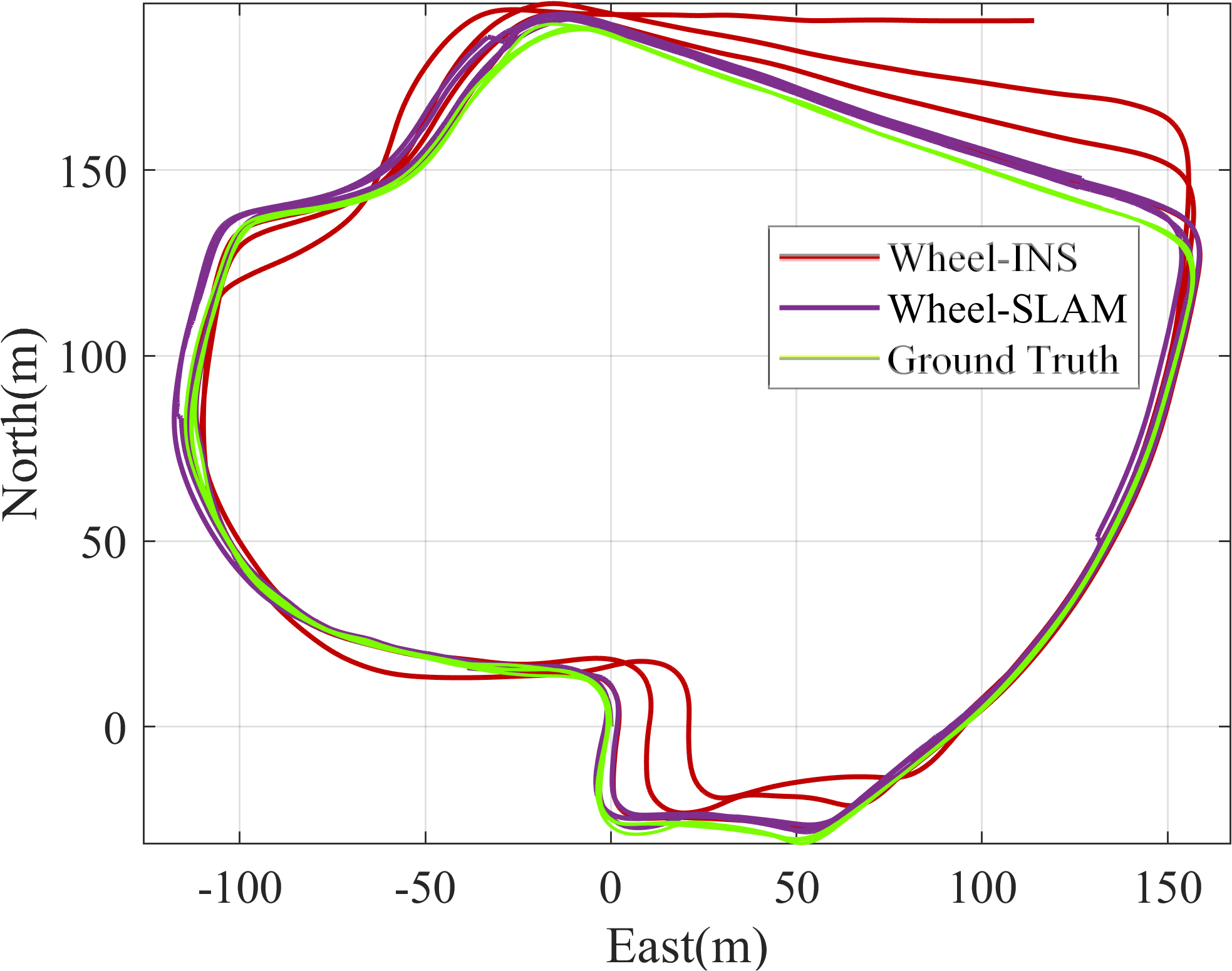}
			}
		\end{minipage}
		\begin{minipage}[t]{0.5\linewidth}
		    \centering
			\subfigure[Horizontal positioning and heading errors in Seq. 3.]{
				\includegraphics[height = 0.7\linewidth]{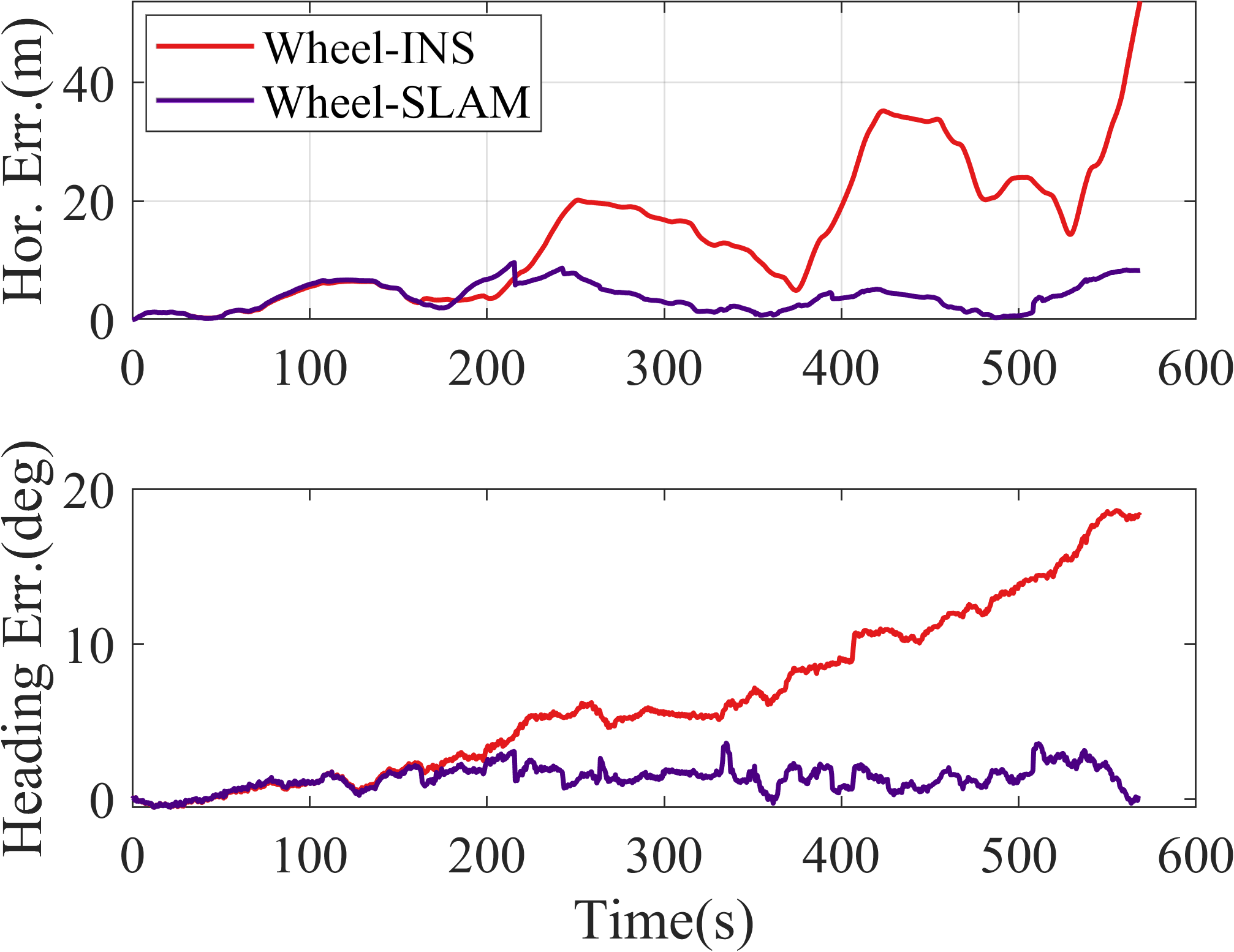}
			}
		\end{minipage}	
	\end{tabular}	
	
	\begin{tabular}{cc}
		\begin{minipage}[t]{0.5\linewidth}
		    \centering
			\subfigure[Estimated trajectories against ground truth in Seq. 4.]{
				\includegraphics[height = 0.7\linewidth]{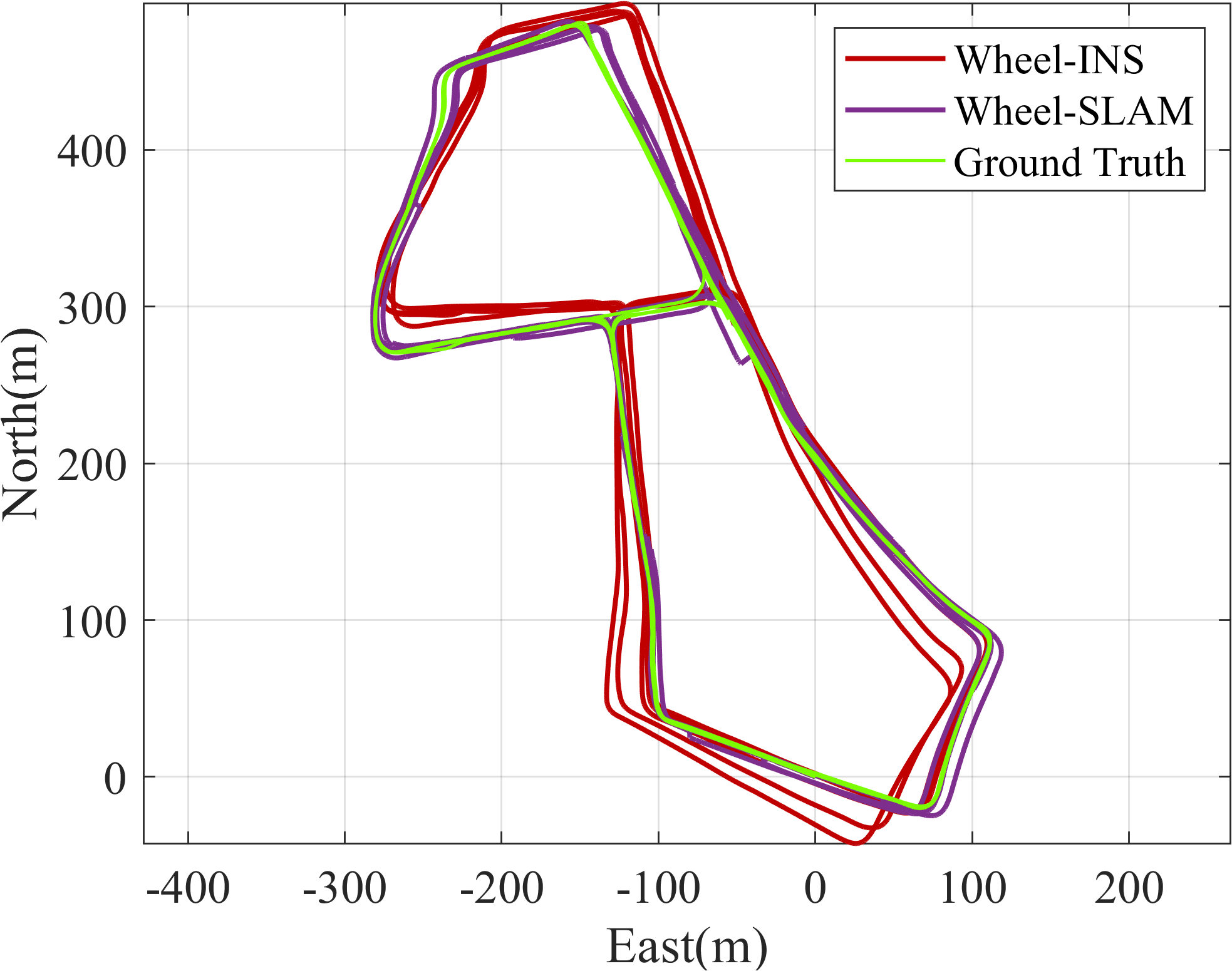}
			}
		\end{minipage}
		\begin{minipage}[t]{0.5\linewidth}
		    \centering
			\subfigure[Horizontal positioning and heading errors in Seq. 4.]{
				\includegraphics[height = 0.7\linewidth]{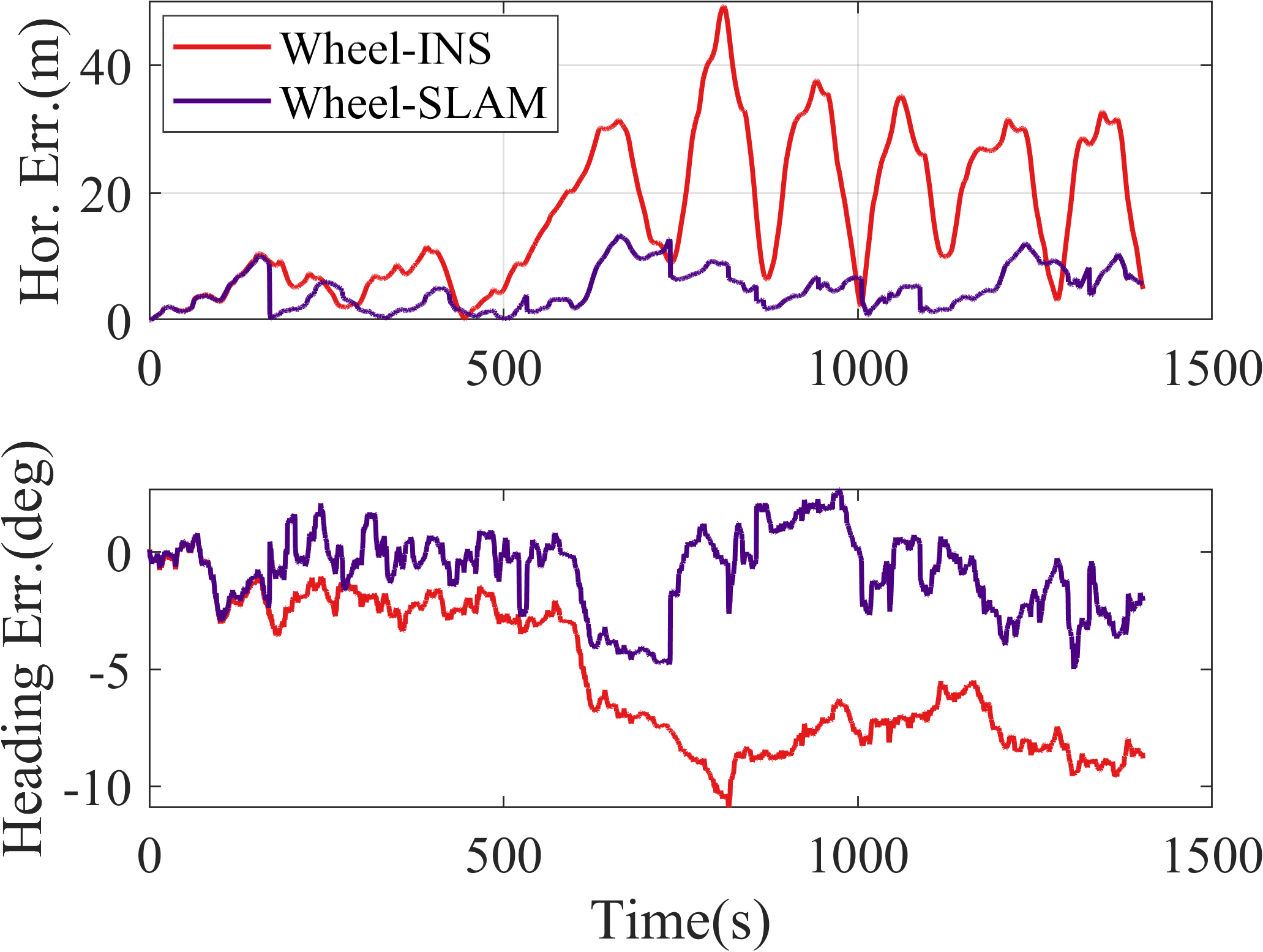}
			}
		\end{minipage}
	\end{tabular}
	
	\begin{tabular}{cc}
		\begin{minipage}[t]{0.5\linewidth}
		    \centering
			\subfigure[Estimated trajectories against ground truth in Seq. 5.]{
				\includegraphics[height = 0.7\linewidth]{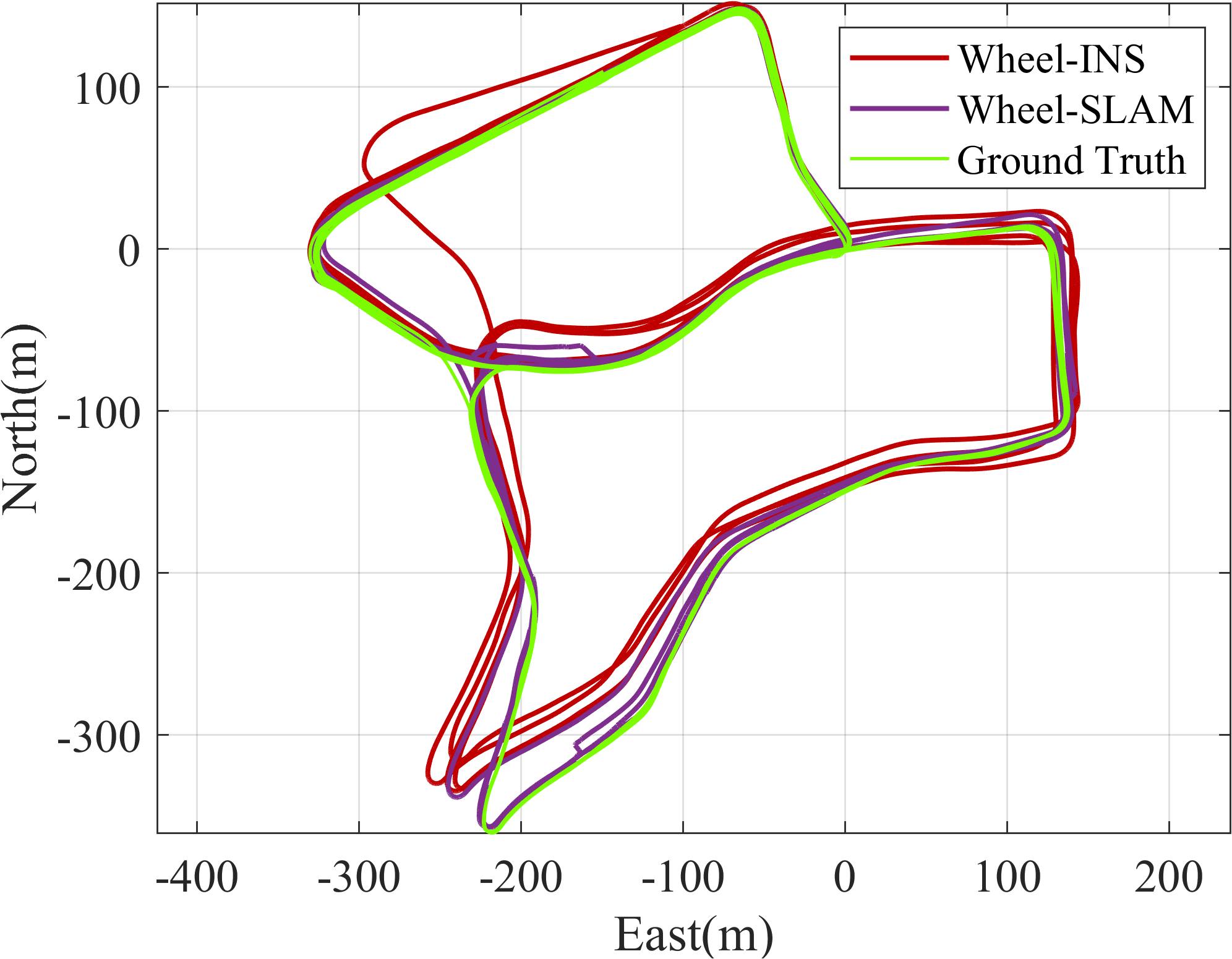}
			}
		\end{minipage}
		\begin{minipage}[t]{0.5\linewidth}
		    \centering
			\subfigure[Horizontal positioning and heading errors in Seq. 5.]{
				\includegraphics[height = 0.7\linewidth]{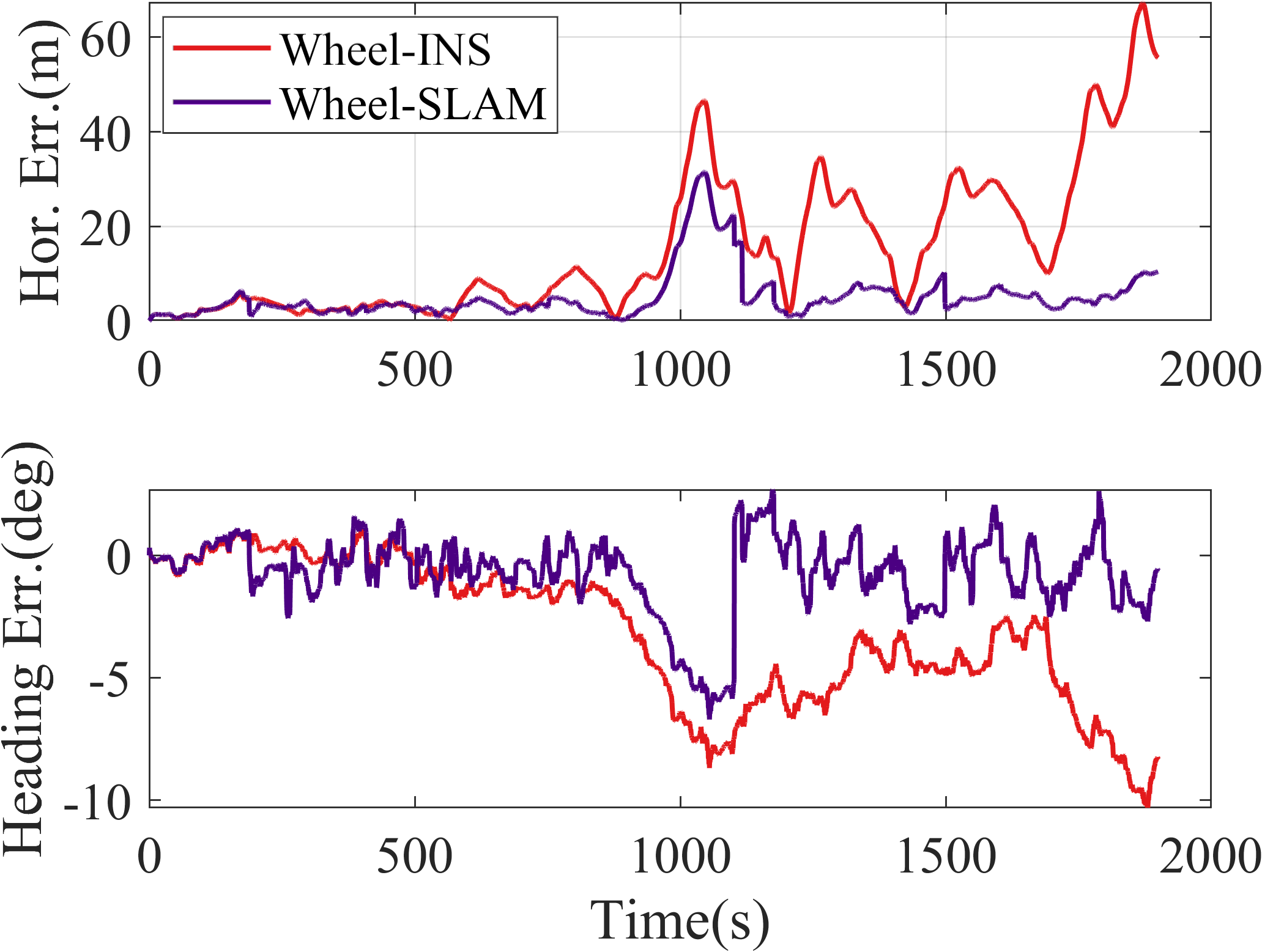}
			}
		\end{minipage}
	\end{tabular}
\caption{The estimated trajectories and corresponding horizontal position and heading errors of Wheel-INS and Wheel-SLAM in the five experiments.}
\label{figtrajcomp2}
	
\end{figure}

\subsection{Performance Comparison and Analysis}

\subsubsection{Performance Comparison}
Fig. \ref{figtrajcomp2} compares the positioning and heading errors of Wheel-SLAM and Wheel-INS in all five experiments, respectively. Please refer to our source code webpage for the positioning results of Wheel-SLAM overlaid on Google earth. It is obvious that compared to Wheel-INS, Wheel-SLAM can suppress the error drift effectively, even in complicated scenarios where the car moves in different directions in large-scale environments, e.g., Seq. 4 and Seq. 5. However, the positioning error is increasing with drift of heading error in Wheel-INS. In addition, we can also notice that there are periodic decreases in the positioning error of Wheel-INS. This is due to the loop closure of the trajectory which cancels part of the cumulative error as discussed in our prior work \cite{niu2021}. 

At the start of the experiments, the positioning results of Wheel-SLAM drift together with Wheel-INS because all the particles are sampled from Wheel-INS according to the Gaussian distribution and there is no external measurement to correct the position error. Once a credible loop closure is reported by some particles, they get a larger weight. Accordingly, the vehicle position would deflect to the previous trajectory when the vehicle visited the same place the last time. It can be observed that there are some sharp decreases in the positioning and heading error of Wheel-SLAM in the experiments, such as the 150\,s in Fig. \ref{figtrajcomp2}(d) and the 1000\,s in Fig. \ref{figtrajcomp2}(j). This is because we use the weighted average value of all the particles as the output. The increased weights of those particles which detect the loop closure would drag the vehicle position to the previous trajectory. In addition, we only correct the current robot state but not the historical trajectory. Using modern graph optimization tools, e.g., gtsam \cite{gtsam}, can jointly optimize the history state so as to make the trajectory smoother and improve the overall accuracy, but this letter mainly focuses on the feasibility to extract road features to enable loop closure in Wheel-INS.

We can observe from the figures that both the positioning and heading error of Wheel-INS increase quickly over time, while the same value of Wheel-SLAM is constrained at a certain level. When there is no loop closure, Wheel-SLAM exhibits the same drift as Wheel-INS in the first circle of the trajectory. However, as the vehicle continues to revisit the roads previously visited, the position drift of Wheel-SLAM can be limited as that in the first circle because of the loop closure mechanism. This suggests that the proposed method is effective to limit the error growth of Wheel-INS by using terrain information perceived by the Wheel-IMU to perform the loop closure.

It is worth mentioning that in Seq. 5, there is a large position and heading drift from about 860\,s to 1100\,s (see Fig. \ref{figtrajcomp2}(j)). This is because the vehicle position error of Wheel-INS drifts quickly at that time. Meanwhile, there is no loop closure reported by the particles, so Wheel-SLAM exhibits a similar drift trend during this time period. Afterward, a loop closure is successfully detected in Wheel-SLAM at about 1100\,s, resulting in a radical error correction.

\begin{table}[t]
	\centering
	\caption{Positioning and Heading Error Statistics of Wheel-SLAM and Wheel-INS}
	\label{TaberrStat}
	\begin{threeparttable}
		\begin{tabular}{p{0.6cm}<{\centering}p{1.3cm}<{\centering}p{1.7cm}<{\centering}p{1.3cm}<{\centering}p{1.7cm}<{\centering}}
			\toprule
			\multirow{2}*{\makecell{Seq.}} &  \multicolumn{2}{c}{Horizontal Pos. RMSE(m)} & \multicolumn{2}{c}{Heading RMSE $(^\circ)$} \\
			\specialrule{0em}{1.5pt}{1.5pt}
			& Wheel-INS & Wheel-SLAM & Wheel-INS & Wheel-SLAM \\
			\midrule
			\specialrule{0em}{1.5pt}{1.5pt}
			{1}&5.70& \textbf{2.50}&1.96&\textbf{1.00}\\
			\specialrule{0em}{1.5pt}{1.5pt}
			{2}& 27.09& \textbf{9.38}&11.36&\textbf{3.17}\\
			\specialrule{0em}{1.5pt}{1.5pt}
			{3}& 18.03& \textbf{8.27}&8.32&\textbf{3.83}\\
			\specialrule{0em}{1.5pt}{1.5pt}
			{4}& 20.24& \textbf{9.21}&6.09&\textbf{2.43}\\
			\specialrule{0em}{1.5pt}{1.5pt}
			{5}&21.44& \textbf{14.42}&4.26&\textbf{2.95}\\
			
			\bottomrule
		\end{tabular}
	\end{threeparttable}
\end{table}

Table \ref{TaberrStat} lists the error statistics of the navigation results of Wheel-SLAM and Wheel-INS in all five experiments. We calculate the root mean square error (RMSE) of the horizontal position and heading (denoted as Horizontal Pos. RMSE and Heading RMSE in \ref{TaberrStat}, respectively) as the indicators to evaluate the navigation performance. Additionally, given the stochasticity of the particle sampling, we run the algorithm multiple times to get the mean value of each test. It can be learned that thanks to the loop closure detection, Wheel-SLAM overwhelmingly outperforms Wheel-INS in terms of both position and heading estimation. Compared to Wheel-INS, the positioning and heading accuracy has been improved by 32.7\%\,$ \sim $\,65.4\% and 30.7\%\,$ \sim $\,72.1\%, respectively. 

\begin{figure*}[t]
	\centering
	\includegraphics[width=15.5cm]{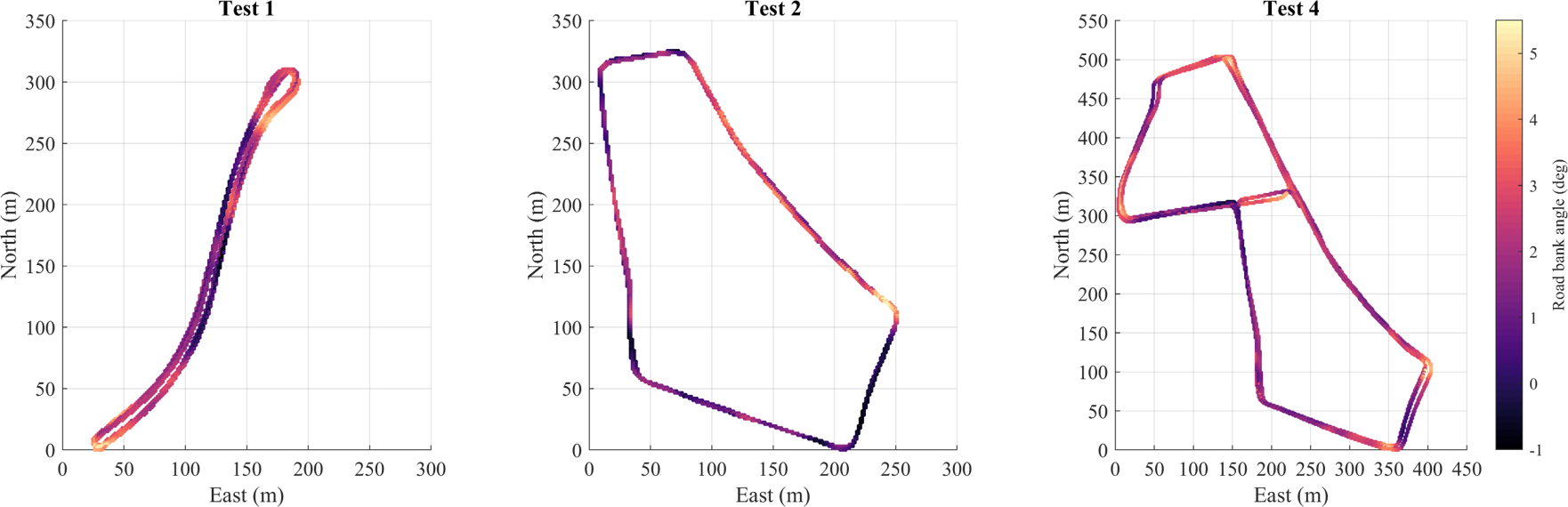}
	\caption{The terrain maps estimated by Wheel-SLAM in Seq. 1, Seq. 2, and Seq. 4, respectively. The colors represent the values of the road bank angles. The larger the road bank angle, the lighter the color.}
	\label{figrollmap}
\end{figure*}

As Seq. 1, Seq. 2, and Seq. 3 are rather simple, there are more opportunities for loop closures. Therefore, the performance improvements in these three tests are more significant than that in Seq. 4 and Seq. 5. 

The terrain maps built by Wheel-SLAM in Seq. 1, Seq. 2, and Seq. 4 are shown in Fig. \ref{figrollmap}. Because the wheel of the vehicle contacts the ground directly, the mapping is not affected by the suspension system of the vehicle especially when the vehicle maneuvering is large which is the case when the IMU is mounted on the vehicle body. Furthermore, these maps can be used to provide valuable information to monitor the deformation and deterioration of the roads. 

\begin{figure}[t]
	\centering
	\includegraphics[width=7cm]{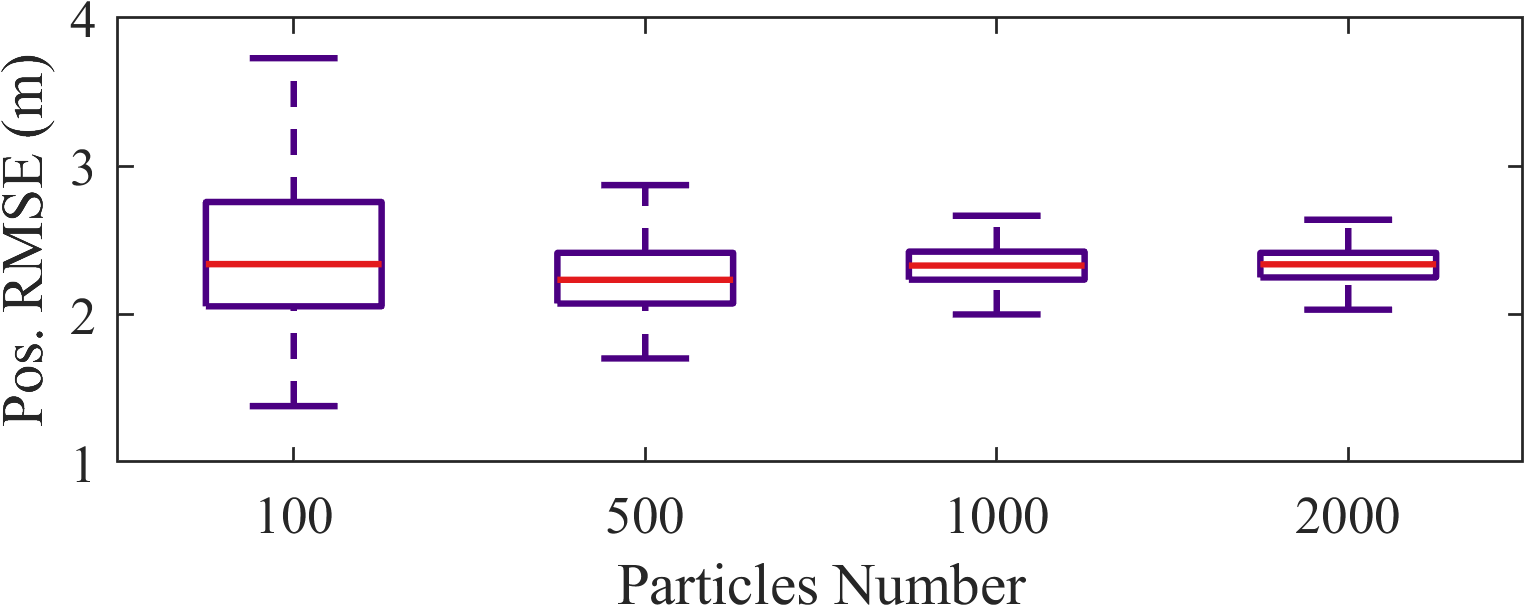}
	\caption{The positioning RMSE of Wheel-SLAM with different particle numbers in Seq. 1. Medians are indicated by the red lines, while the bottom and top edges of the boxes indicate the first quartile and third quartile, respectively, and the whiskers show the maximum (upper) and minimum (lower).}
	\label{figparticlesComp}
\end{figure}

\subsubsection{Analysis on the characteristics of Wheel-SLAM}
To further evaluate the performance and stability of Wheel-SLAM, we set different particle numbers to compare the positioning performance. The algorithm was run 100 times for each configuration. Fig. \ref{figparticlesComp} shows the results. 

It can be observed that the position error of Wheel-SLAM is more centralized when there are more particles, which means that the stability of Wheel-SLAM is improved with the increment of particles. This is because it is more likely for the system to detect the real loop closure with more particles and the performance would also be less susceptible to anomalies. However, there is no obvious gain by adding particles from 1000 to 2000 because of the diminishing marginal effect. In addition, the accuracy of Wheel-SLAM also depends on the position error when the vehicle visits the place for the first time. In consequence, continuing to add the particles only improve position accuracy lightly. 

Furthermore, it can be learned that in a statistical sense, the overall positioning performance of Wheel-SLAM is also not significantly enhanced with the increase of the particles. Although there are some outliers when the particle number is small (= 100), the system is robust to recognize the loop closure thanks to the excellent DR ability of Wheel-INS.   

\subsection{Discussion}
The core principles behind Wheel-SLAM can be summarized as follows: 1) Particles are spread to sample the possible state of the robot and detect the loop closure by the trajectory maintained by each particle; 2) The road bank angle sequence matching result is used to update the particle weights, so as to pick out the most trusted one(s). What plays a central role in Wheel-SLAM is the roll sequence matching strategy. It must be robust enough to keep the outstanding particles while filtering out false alarms. Therefore, we used a rather strict loop closure detection criteria to make the loop closure detection robust.


However, it can be realized that there are two major limitations in the application of Wheel-SLAM. First, the robot must strictly revisit the previous places with a certain length. It is not like the vision-based SLAM where the vehicle has the remote sensing ability by using exteroceptive sensors, e.g., camera and LiDAR. In Wheel-SLAM, the Wheel-IMU is used to extract the terrain features which can only be obtained by the exact arrival of the robot. Second, the success of the loop closure depends on the matching of the road bank angle sequence. If the robot is moving on extremely smooth roads without any fluctuations in the bank angle, it would be difficult to detect loop closure.   

\section{CONCLUSIONS}
In this study, we propose to perform SLAM with only one Wheel-IMU by exploiting the environmental perception ability of the Wheel-IMU. To be specific, we extend our previous study on Wheel-INS to Wheel-SLAM by extracting the terrain feature from the robot roll angle estimates to enable loop closure detection. The system is implemented with a Rao-Blackwellized particle filter where each particle maintains its own robot state and grid map.
Experimental results show that the proposed method can effectively suppress the error drift of Wheel-INS. The positioning and heading accuracy has been averagely improved by 52.6\% and 53.2\%, respectively, against Wheel-INS.

However, Wheel-SLAM has two major limitations. First, a certain level of variation in the bank angles of the road is needed. 
Second, the robot has to revisit the same place exactly. 
Wheel-SLAM would be suitable for those kinds of robots that move repeatedly in given areas, for example, the sweeping robots and patrol robots in restricted areas.

For future research, integrating Wheel-SLAM with other exteroceptive sensors (e.g., camera and LiDAR) would be promising to improve the robustness and applicability of the robot navigation system.



\bibliographystyle{IEEEtranTIE}
\bibliography{ReferenceWheel-SLAM_2022}

\begin{thebibliography}{10}
\providecommand{\url}[1]{#1}
\csname url@samestyle\endcsname
\providecommand{\newblock}{\relax}
\providecommand{\bibinfo}[2]{#2}
\providecommand{\BIBentrySTDinterwordspacing}{\spaceskip=0pt\relax}
\providecommand{\BIBentryALTinterwordstretchfactor}{4}
\providecommand{\BIBentryALTinterwordspacing}{\spaceskip=\fontdimen2\font plus
\BIBentryALTinterwordstretchfactor\fontdimen3\font minus
  \fontdimen4\font\relax}
\providecommand{\BIBforeignlanguage}[2]{{%
\expandafter\ifx\csname l@#1\endcsname\relax
\typeout{** WARNING: IEEEtran.bst: No hyphenation pattern has been}%
\typeout{** loaded for the language `#1'. Using the pattern for}%
\typeout{** the default language instead.}%
\else
\language=\csname l@#1\endcsname
\fi
#2}}
\providecommand{\BIBdecl}{\relax}
\BIBdecl

\bibitem{niu2021}
X.~Niu, Y.~Wu, and J.~Kuang, ``{Wheel-INS}: A wheel-mounted {MEMS} {IMU}-based
  dead reckoning system,'' \emph{{IEEE} Trans. Veh. Technol.}, vol.~70, no.~10,
  pp. 9814--9825, 2021.

\bibitem{wu2021}
Y.~Wu, X.~Niu, and J.~Kuang, ``A comparison of three measurement models for the
  wheel-mounted {MEMS} {IMU}-based dead reckoning system,'' \emph{{IEEE} Trans.
  Veh. Technol.}, vol.~70, no.~11, pp. 11\,193--11\,203, 2021.

\bibitem{liyou2021}
N.~El-Sheimy and Y.~Li, ``Indoor navigation: State of the art and future
  trends,'' \emph{Satell. Navig.}, vol.~2, no.~1, pp. 1--23, 2021.

\bibitem{qin2018}
T.~Qin, P.~Li, and S.~Shen, ``{VINS-Mono}: A robust and versatile monocular
  visual-inertial state estimator,'' \emph{{IEEE} Trans. Robot.}, vol.~34,
  no.~4, pp. 1004--1020, 2018.

\bibitem{huang2019}
G.~Huang, ``Visual-inertial navigation: A concise review,'' in \emph{Proc. Int.
  Conf. Robot. Automat.}, pp. 9572--9582.\hskip 1em plus 0.5em minus
  0.4em\relax IEEE, 2019.

\bibitem{xu2022TRO}
W.~Xu, Y.~Cai, D.~He, J.~Lin, and F.~Zhang, ``{FAST-LIO2}: Fast direct
  lidar-inertial odometry,'' \emph{{IEEE} Trans. Robot.}, vol.~38, no.~4, pp.
  2053--2073, 2022.

\bibitem{wu2019sensors}
Y.~Wu, X.~Niu, J.~Du, L.~Chang, H.~Tang, and H.~Zhang, ``Artificial marker and
  mems imu-based pose estimation method to meet multirotor uav landing
  requirements,'' \emph{Sensors}, vol.~19, no.~24, p. 5428, 2019.

\bibitem{ouyang2020}
W.~Ouyang, Y.~Wu, and H.~Chen, ``{INS}/odometer land navigation by accurate
  measurement modeling and multiple-model adaptive estimation,'' \emph{{IEEE}
  Trans. Aerosp. Electron. Syst.}, vol.~57, no.~1, pp. 245--262, 2021.

\bibitem{slamreview2016}
C.~Cadena, L.~Carlone, H.~Carrillo, Y.~Latif, D.~Scaramuzza, J.~Neira, I.~Reid,
  and J.~J. Leonard, ``Past, present, and future of simultaneous localization
  and mapping: Toward the robust-perception age,'' \emph{{IEEE} Trans. Robot.},
  vol.~32, no.~6, pp. 1309--1332, 2016.

\bibitem{litianyi2019}
T.~Li, M.~Yang, H.~Li, L.~Deng, and C.~Wang, ``A terrain-based vehicle
  localization approach robust to braking,'' \emph{{IEEE} Trans. Intell.
  Transp. Syst.}, vol.~20, no.~8, pp. 2923--2932, 2019.

\bibitem{Zhang2020rs}
H.~Zhang, W.~Li, C.~Qian, and B.~Li, ``A real time localization system for
  vehicles using terrain-based time series subsequence matching,'' \emph{Remote
  Sensing}, vol.~12, no.~16, p. 2607, 2020.

\bibitem{Gim2021Access}
J.~Gim and C.~Ahn, ``Inertial-sensor-based non-dead reckoning localization for
  ground vehicles,'' \emph{IEEE Access}, vol.~9, pp. 132\,766--132\,778, 2021.

\bibitem{Vemulapalli2011ACC}
P.~K. Vemulapalli, A.~J. Dean, and S.~N. Brennan, ``Pitch based vehicle
  localization using time series subsequence matching with multi-scale extrema
  features,'' in \emph{Proc. Amer. Control Conf.}, pp. 2405--2410, 2011.

\bibitem{Emil2015TITS}
E.~I. Laftchiev, C.~M. Lagoa, and S.~N. Brennan, ``Vehicle localization using
  in-vehicle pitch data and dynamical models,'' \emph{{IEEE} Trans. Intell.
  Transp. Syst.}, vol.~16, no.~1, pp. 206--220, 2015.

\bibitem{Ahmed2016VTC}
H.~Ahmed and M.~Tahir, ``Terrain-based vehicle localization using low cost
  mems-imu sensors,'' in \emph{Proc. IEEE Veh. Technol. Conf.}, pp. 1--5, 2016.

\bibitem{Dean2008PhD}
A.~J. Dean, ``Terrain-based road vehicle localization using attitude
  measurements,'' Ph.D. dissertation, Department of Mechanical and Nuclear
  Engineering, The Pennsylvania State University, Philadelphia, U.S., 2008.

\bibitem{Ryu2004ACC}
J.~Ryu and J.~Gerdes, ``Estimation of vehicle roll and road bank angle,'' in
  \emph{Proc. Amer. Control Conf.}, vol.~3, pp. 2110--2115, 2004.

\bibitem{Dean2011VSD}
A.~Dean, R.~Martini, and S.~Brennan, ``Terrain-based road vehicle localisation
  using particle filters,'' \emph{Veh. Syst. Dyn.}, vol.~49, no.~8, pp.
  1209--1223, 2011.

\bibitem{Martini2006PhD}
R.~D. Martini, ``{GPS}/{INS} sensing coordination for vehicle state
  identification and road grade positioning,'' Master's thesis, Department of
  Mechanical and Nuclear Engineering, The Pennsylvania State University,
  Philadelphia, U.S., 2006.

\bibitem{chen2018aaai}
C.~Chen, X.~Lu, A.~Markham, and N.~Trigoni, ``Ionet: Learning to cure the curse
  of drift in inertial odometry,'' \emph{Proc. AAAI Conf. on Artif. Intell.},
  vol.~32, no.~1, 2018.

\bibitem{herath2022CVPR}
S.~Herath, D.~Caruso, C.~Liu, Y.~Chen, and Y.~Furukawa, ``Neural inertial
  localization,'' in \emph{Proceedings of the IEEE/CVF Conf. on Computer Vision
  and Pattern Recognition}, pp. 6604--6613, Jun. 2022.

\bibitem{martin2020tiv}
M.~Brossard, A.~Barrau, and S.~Bonnabel, ``Ai-imu dead-reckoning,'' \emph{IEEE
  Trans. on Intelli. Veh.}, vol.~5, no.~4, pp. 585--595, 2020.

\bibitem{Angermann2012IEEEProc}
M.~Angermann and P.~Robertson, ``{FootSLAM}: Pedestrian simultaneous
  localization and mapping without exteroceptive sensors-hitchhiking on human
  perception and cognition,'' \emph{Proc. {IEEE}}, vol. 100, pp. 1840--1848,
  2012.

\bibitem{dissanayake2001}
G.~Dissanayake, S.~Sukkarieh, E.~Nebot, and H.~Durrant-Whyte, ``The aiding of a
  low-cost strapdown inertial measurement unit using vehicle model constraints
  for land vehicle applications,'' \emph{{IEEE} Trans. Robot.}, vol.~17, no.~5,
  pp. 731--747, 2001.

\bibitem{thrun2005probabilistic}
S.~Thrun, W.~Burgard, and D.~Fox, \emph{Probabilistic Robotics}.\hskip 1em plus
  0.5em minus 0.4em\relax MIT Press, 2005.

\bibitem{Djuric2003IEEESPM}
P.~Djuric, J.~Kotecha, J.~Zhang, Y.~Huang, T.~Ghirmai, M.~Bugallo, and
  J.~Miguez, ``Particle filtering,'' \emph{{IEEE} Signal Process. Mag.},
  vol.~20, no.~5, pp. 19--38, 2003.

\bibitem{murphy2001rao}
K.~Murphy and S.~Russell, ``{Rao-Blackwellised} particle filtering for dynamic
  {Bayesian} networks,'' in \emph{Sequential Monte Carlo methods in practice},
  pp. 499--515.\hskip 1em plus 0.5em minus 0.4em\relax Springer, 2001.

\bibitem{cyrill2007TRO}
G.~Grisetti, C.~Stachniss, and W.~Burgard, ``Improved techniques for grid
  mapping with {Rao-Blackwellized} particle filters,'' \emph{{IEEE} Trans.
  Robot.}, vol.~23, no.~1, pp. 34--46, 2007.

\bibitem{Montemerlo2002AAAI}
M.~Montemerlo, S.~Thrun, D.~Koller, B.~Wegbreit \emph{et~al.}, ``{FastSLAM}: A
  factored solution to the simultaneous localization and mapping problem,'' in
  \emph{Proc. AAAI Conf. on Artif. Intell.}, pp. 593--598, 2002.

\bibitem{gtsam}
\BIBentryALTinterwordspacing
F.~Dellaert, R.~Roberts, V.~Agrawal, and et~al, ``borglab/gtsam,'' May. 2022.
  [Online]. Available: \url{https://doi.org/10.5281/zenodo.5794541}
\BIBentrySTDinterwordspacing

\end{thebibliography}

\end{document}